\RequirePackage[T1]{fontenc}
\documentclass[letterpaper,10 pt,conference]{ieeeconf}
\IEEEoverridecommandlockouts
\usepackage{graphicx}
\usepackage{amsmath,amssymb}
\usepackage{booktabs}
\usepackage{algorithm}
\usepackage{algorithmic}
\usepackage[hidelinks]{hyperref}
\hypersetup{pdftitle={Characterizing Replay Retention Under Dynamics Shift in Model-Based Reinforcement Learning},pdfauthor={Everest Yang, Skye Thompson, George D. Konidaris}}
\newcommand{\corr}{\mathrm{AUC}_{\mathrm{age}}}
\newcommand{\FreshGain}{+68\,\pm\,26}

\newcommand{\FreshMild}{-154\,\pm\,36}

\newcommand{\ResetRecurringGap}{175\,\pm\,32}
\graphicspath{{figures/}}
\title{\LARGE\bf Characterizing Replay Retention Under Dynamics Shift\\
in Model-Based Reinforcement Learning}
\author{Everest Yang$^{*}$, Skye Thompson, George D. Konidaris\\
Brown University, Providence, RI, USA%
\thanks{$^{*}$Correspondence to everest\_yang@brown.edu}}
\begin{document}
\clubpenalty=10000
\widowpenalty=10000
\raggedbottom
\maketitle
\thispagestyle{empty}
\pagestyle{empty}

\begin{abstract}
Adapting to changes in robot dynamics requires learning from new data without discarding experience that may still be useful. In continual model-based reinforcement learning (RL), replay collected before a dynamics change can slow adaptation, while removing it unnecessarily reduces available training data and can be especially costly if earlier dynamics return. We study when recent transitions are preferable to the full replay history. Two quantities characterize this trade-off: change magnitude and age--staleness area under the curve (AUC), measuring how well transition age separates stale from fresh data. Forgetting stale data helps after large permanent shifts but hurts when dynamics recur and older data becomes useful again. Choosing a replay strategy therefore depends on predicting when older data will help or hurt. We test these effects across two locomotion morphologies, two model-based RL algorithms, and Real-World RL benchmark perturbations. Because ground-truth staleness labels are unavailable on deployed robots, we evaluate whether an estimator built from interaction data can still provide the quantities needed to choose a replay strategy after permanent changes. Our results show that replay retention depends on change magnitude and on how the dynamics evolve.
\end{abstract}

\section{Introduction}
Reinforcement learning (RL) with learned world models is a common approach to continuous control when an agent must adapt as its dynamics change. Methods such as DreamerV3~\cite{hafner2023dreamerv3} and TD-MPC2~\cite{hansen2024tdmpc2} learn from replayed experience~\cite{lin1992replay} to predict the consequences of actions. Adapting a learned world model can become difficult when the robot's dynamics shift during continued training, for example from actuator damage or a change in payload. Older transitions may then no longer reflect the robot's current behavior. Training on outdated transitions can slow down adaptation, while removing them may discard data that could still be useful under the current dynamics.

\begin{figure*}[t]
\centering
\includegraphics[width=\textwidth]{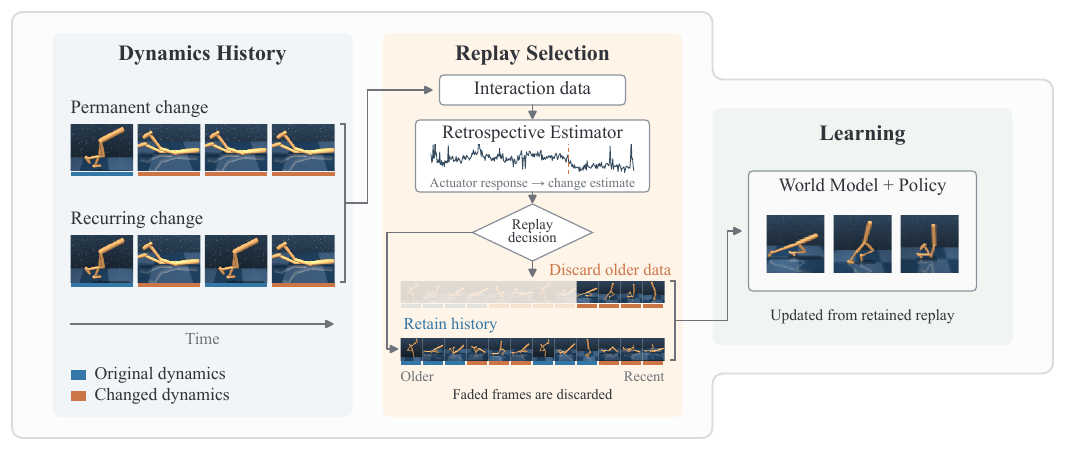}
\caption{Overview of replay retention under changing dynamics. Interaction data populate replay and can also be used retrospectively to estimate whether older transitions should be retained or discarded. The resulting replay choice determines which data would be used to update the world model and policy. Under permanent changes, older data may remain mismatched, while under recurring dynamics it can become useful again.}
\label{fig:hero}
\end{figure*}

The value of older replay data depends on how the dynamics evolve over time (Fig.~\ref{fig:hero}). After a lasting change, such as a new payload, transitions collected beforehand can remain mismatched. If earlier dynamics later return, those same transitions can become useful again. The utility of older replay data therefore depends on both the size and temporal structure of the dynamics change.

We study how transition staleness affects replay usefulness by comparing replay over the full history with fixed windows of recent data. We focus on two properties of the dynamics change. The first is its magnitude, which captures how much the underlying dynamics change relative to those represented by older transitions. The second is the relationship between transition age and continued relevance, which we measure using age--staleness AUC. Age is useful for replay selection when older transitions are more likely to come from outdated dynamics, but it becomes less informative when the dynamics recur and older transitions once again match the dynamics the agent is experiencing.

We study three dynamics patterns: permanent changes, recurring changes, and a no-change control. Within these settings, we vary change magnitude and recurrence. Separate controls vary the size of the replay window and the total number of transitions the replay buffer can store, testing how each affects performance across these dynamics settings. We test these effects across two locomotion morphologies, two model-based RL algorithms, and perturbations from the Real-World RL (RWRL) benchmark. We then ask whether interaction data can provide enough information to decide when older replay should be discarded. For permanent changes, we also show that we can estimate change magnitude and age--staleness from actuator response alone, without access to the simulator's dynamics parameters, suggesting that these signals could guide replay selection on real-world systems. Our contributions are as follows:
\begin{enumerate}
\item A characterization of when older replay data should be retained based on change magnitude and the relationship between transition age and staleness.
\item Evidence that the same dynamics change can favor forgetting when it persists and retention when it recurs. This shows that change magnitude alone is not enough to select an effective replay strategy.
\item A retrospective actuator-response analysis that detects dynamics changes and decides whether older replay should be discarded using estimated change magnitude and age--staleness, with staleness inferred from observed actuator responses rather than ground-truth dynamics parameters.
\end{enumerate}

\section{Related Work}
When dynamics change during continued learning, adaptation depends on how the model is updated and which past transitions remain available for replay. Prior studies have examined replay retention and sampling, including the effects of transition age and buffer size on control~\cite{debruin2018experience,zhang2017deeper,fedus2020replay}. Local forgetting removes samples near parts of the state space that have changed so a model-based agent can adapt without discarding data elsewhere~\cite{rahimi2023localforgetting}. Curious Replay instead prioritizes transitions that are still informative for improving the world model~\cite{kauvar2023curious}, and Continual-Dreamer studies selective replay as a way to reduce forgetting while preserving transfer across tasks~\cite{kessler2023continualdreamer}. World Models with Augmented Replay (WMAR) maintains a more diverse long-term replay distribution~\cite{yang2024wmar}. Our work focuses on when age-based forgetting is useful under permanent versus recurring dynamics changes. More selective replay strategies can avoid some of the losses from a fixed window of recent data. We focus on the separate question of when transition age is useful for deciding which transitions to retain.

Another approach is to adapt the model or policy directly rather than modify replay. Context-aware dynamics models infer latent information about the current dynamics~\cite{lee2020cadm}. Rapid motor adaptation instead uses recent interaction history to adapt locomotion policies online~\cite{kumar2021rma}. Meta-learned dynamics models can adapt from recent data after changes in morphology, terrain, or payload~\cite{nagabandi2019learningtoadapt}. Dynamics randomization exposes policies to variations in physical parameters during training to improve robustness~\cite{peng2018dynamicsrand}, while other meta-RL methods learn priors or latent representations that support rapid adaptation~\cite{finn2017maml,rakelly2019pearl}. Cully et al. use a map of high-performing behaviors to recover from robot damage~\cite{cully2015robots}. Our setting instead keeps the learning procedure fixed and studies which previously collected transitions should remain available as training continues. The same model continues learning as the dynamics change, rather than relying on domain randomization or a separate model for each regime.

Prioritized replay samples transitions according to temporal-difference error~\cite{schaul2016prioritized}. Density-ratio importance weighting favors transitions that better match the current policy distribution~\cite{sinha2022lfiw}. Methods with adaptive windows choose the window size from the data stream~\cite{bifet2007adwin}.

\section{Problem Formulation}
\label{sec:problem}
\subsection{Continual Learning and Retention}
We consider a model-based RL agent that continues updating its world model and policy after the environment dynamics change~\cite{khetarpal2022continual}. At each step, the agent acts, stores the resulting transition, and samples from replay for further updates, while the learned model itself is never reset. The replay strategy determines which stored transitions remain available for training.

\textbf{Passive} replay samples from the full available buffer, with a capacity of one million transitions. \textbf{Recency} restricts replay to the newest $w=10{,}000$ transitions. Let $J_s$ denote the mean episodic return for replay strategy $s$, averaged over evaluations during the 90k frames following the dynamics change. Each evaluation uses ten episodes. Our primary comparison is the paired difference in return after the dynamics shift
\begin{equation}
D = J_{\mathrm{recency}} - J_{\mathrm{passive}}.
\end{equation}
$D>0$ favors Recency. We call $D$ the Recency advantage.

\subsection{Staleness Labels and Age Ranking}
In our controlled experiments, the dynamics take one of two known regimes. Each transition is labeled by the dynamics regime in which it was collected, such as the original dynamics or a shifted regime with reduced actuator gain. Let $r_t$ denote the regime active at time $t$. A transition collected at time $\tau$ is stale at time $t$ when $r_\tau \neq r_t$. Here, staleness is defined from the known dynamics regime, giving us ground-truth staleness labels for the controlled experiments.

Age-based forgetting is useful only when transition age is informative about staleness. We measure how well transition age $A_t(\tau)=t-\tau$ separates stale from fresh data using the area under the receiver operating characteristic curve (AUC), which we define for a buffer at time $t$ as
\begin{align}
\corr(t)={}&\Pr[A_t(x_{\mathrm{stale}})>A_t(x_{\mathrm{fresh}})]\nonumber\\
&+\tfrac12\Pr[A_t(x_{\mathrm{stale}})=A_t(x_{\mathrm{fresh}})].
\label{eq:auc}
\end{align}
This is the probability that a random stale transition is older than a random fresh one~\cite{hanley1982auc}. An AUC of 1 means stale transitions are always older than fresh ones, while 0.5 corresponds to chance ranking. The measure is undefined when the buffer contains only stale or only fresh transitions.

Under recurring dynamics, a transition can be stale at one point in training and useful again later when its dynamics return. Because this relationship changes over time, we report age--staleness AUC averaged over the evaluation window after the change. Each episode is assigned to the dynamics regime active when that episode ends.

\section{Characterizing Replay Retention Under Dynamics Shift}
\label{sec:approach}
\subsection{Bias--Variance Trade-offs in Replay Retention}
Consider estimating a scalar component of the current dynamics from $n_o$ observations of the old regime and $n_f$ fresh observations, each with noise variance $\sigma^2$. Let the old observations be offset from the current parameter by $M$. If the old data are weighted by $\beta$, the resulting mean-squared estimation error is
\begin{equation}
\mathcal{E}(\beta)=\beta^2 M^2+\beta^2\frac{\sigma^2}{n_o}+(1-\beta)^2\frac{\sigma^2}{n_f}.
\end{equation}
The optimal weight on old data is
\begin{equation}
\beta^\star=\frac{\sigma^2/n_f}{M^2+\sigma^2/n_o+\sigma^2/n_f}.
\end{equation}
When the mismatch is small, retaining old data mainly reduces variance. As $M$ grows, the optimal weight on old data falls toward zero, so larger changes favor forgetting. For a fixed replay window, the benefit of forgetting also depends on how much stale data remains inside that window. Let $\rho$ denote the stale fraction in the full buffer and $\rho_w$ the stale fraction in a window of size $w$. Relative to training on all $N$ observations in the full buffer, restricting training to the window changes the mean-squared estimation error by
\begin{equation}
\Delta\mathcal{E}=(\rho^2-\rho_w^2)M^2-\sigma^2\left(\frac{1}{w}-\frac{1}{N}\right).
\label{eq:risk}
\end{equation}
The first term captures the reduction in bias from removing stale data, while the second captures the variance cost of using fewer samples. Forgetting is beneficial when the former outweighs the latter. Related analyses of moving windows after a change capture the same bias--variance trade-off between retaining stale data and reducing the number of training samples~\cite{kuncheva2009window}. For $\rho>\rho_w$, the error difference is zero at the mismatch threshold
\begin{equation}
M^\star=\sigma\sqrt{\frac{1/w-1/N}{\rho^2-\rho_w^2}}.
\label{eq:mstar}
\end{equation}
Recency is useful only when removing stale data outweighs the cost of training on fewer samples. Small dynamics changes leave older transitions relatively useful, while low age--staleness separation means a window of recent data discards useful experience along with stale data. Age--staleness AUC does not determine $\rho_w$ directly, but it measures how well age distinguishes stale from fresh transitions.

We vary change magnitude, recurrence, window size, and buffer capacity separately to test how the performance trade-off between retaining potentially stale data and training on fewer samples changes across these conditions. The model abstracts away policy updates, nonlinear function approximation, and the changing data distribution during training. Its role is to motivate when removing stale data should outweigh the cost of training on fewer samples, which we test in the experiments that follow.

\begin{figure*}[t]
\centering
\includegraphics[width=\textwidth]{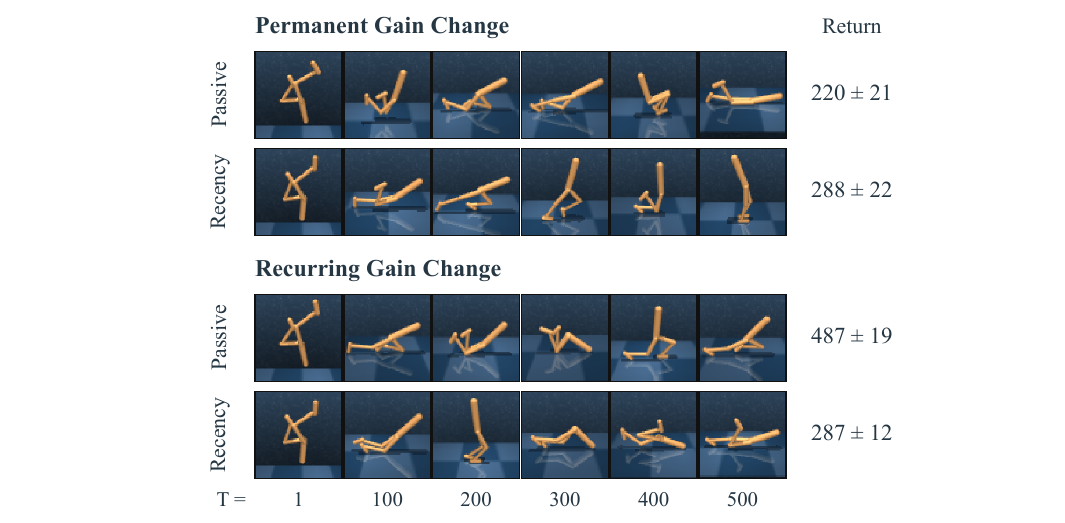}
\caption{Passive and Recency trajectories under permanent and recurring gain changes. Frames are taken from matched training episodes at the indicated steps. Values at right report evaluation return after the shift over five seeds, mean $\pm$ standard error (s.e.).}
\label{fig:rollouts}
\end{figure*}
\subsection{Estimating Regime Changes from Actuator Response}
\label{sec:estimator}
\begin{algorithm}[t]
\caption{Retrospective Replay Selection}
\label{alg:selector}
\begin{algorithmic}[1]
\STATE \textbf{Input:} Interaction trajectory $\mathcal{H}$
\STATE \textbf{Output:} Replay strategy $S$
\STATE $K \leftarrow \textsc{EstimateActuatorResponse}(\mathcal{H})$
\STATE $\hat{\mathbf{s}} \leftarrow \textsc{DetectStaleSegments}(K)$
\STATE $\widehat{\mathrm{AUC}}_{\mathrm{age}} \leftarrow \textsc{AgeStalenessAUC}(\mathcal{H},\hat{\mathbf{s}})$
\STATE $\hat m \leftarrow \textsc{EstimateMagnitude}(K,\hat{\mathbf{s}})$
\IF{$\widehat{\mathrm{AUC}}_{\mathrm{age}}\ge0.87$ \AND $\hat m\ge0.40$}
    \STATE $S \leftarrow \textsc{Recency}$
\ELSE
    \STATE $S \leftarrow \textsc{Passive}$
\ENDIF
\STATE \textbf{return} $S$
\end{algorithmic}
\end{algorithm}
In simulation, we first evaluate replay retention using ground-truth regime labels to measure staleness directly. Outside simulation, those labels would be unavailable, so we separately test whether estimates from the robot's own interaction data can support similar replay decisions without privileged dynamics information. We estimate dynamics changes from the agent's actions and measured joint velocities. For each episode $e$ and joint $j$, we compute
\begin{equation}
K_{e,j}=\frac{\sum_t (v_{j,t+1}-v_{j,t})\,a_{j,t}}{\sum_t a_{j,t}^2+\epsilon}.
\label{eq:kresp}
\end{equation}
$K_{e,j}$ measures actuator response rather than physical gain, since it can also vary with contacts, state visitation, and policy behavior. We evaluate the mean response across joints, a variant restricted to large actions, $|a_{j,t}|>0.5$, and the full vector of per-joint responses.

Each response channel is standardized using its median and median absolute deviation. Pruned Exact Linear Time (PELT)~\cite{killick2012pelt,truong2020ruptures} then segments the episode-level series using squared-error cost, a minimum segment length of five episodes, and penalty $3d\log n$, where $d$ is the number of channels and $n$ the number of episodes. We use change-point detection as a tool for estimating when the interaction dynamics shift, rather than as a replay strategy itself~\cite{alegre2021mbcd,zollicoffer2025novelty}. Earlier segments are labeled stale when their difference from the final segment gives a two-sample $z$-statistic greater than three, a threshold set before the evaluation and not tuned on these runs. These labels provide an estimated AUC at the point where the replay decision is made, and the change magnitude is then estimated from the medians of the stale and fresh segments as
\begin{equation}
\hat m = 1-\frac{\operatorname{median}(K_{\mathrm{fresh}})}{\operatorname{median}(K_{\mathrm{stale}})}.
\label{eq:mhat}
\end{equation}

The replay decision uses two estimated quantities: age--staleness AUC and change magnitude. We set the AUC threshold to $0.87$, corresponding to the sampled recurrence condition where the Recency advantage becomes uncertain for the main 10k replay window. The threshold is not recalibrated for the 20k window, and the selector is evaluated only against the 10k window. We set the magnitude threshold to $0.40$, between the Walker sweep points where Recency hurt ($g=0.7$) and helped ($g=0.5$), and fixed it before the $g=0.6$ run that later placed the sign change between magnitudes $0.4$ and $0.5$ (Section~\ref{sec:sweep}). The selector chooses Recency, meaning it discards older replay, only when both thresholds are exceeded. Algorithm~\ref{alg:selector} summarizes the selector. Because regime segmentation uses the completed trajectory, the current selector is retrospective. An online version could apply the same estimates to the trajectory observed so far, making reliable online change detection a direction for future work.

\section{Experiments}
\label{sec:experiments}
\subsection{Setup and Dynamics Changes}
We use proprioceptive observations from the DeepMind Control Suite (DMC)~\cite{tassa2018dmc}. Walker with DreamerV3 is our main setting, where we study the effects of forgetting under permanent and recurring dynamics changes while varying change magnitude, recurrence period, window size, and buffer capacity. We use Cheetah to test whether the same replay trends hold across different agent morphologies, and TD-MPC2 to test whether they persist with a different model-based RL algorithm. DreamerV3 uses action repeat two and 500-transition episodes, so the $10{,}000$-transition recency window spans about twenty episodes. Each branch runs for 130k environment frames. The dynamics change begins after 40k frames, and we evaluate return over the next 90k frames.

We study three controlled dynamics changes: a permanent gain reduction, permanent actuator damage, and a recurring gain change. We also run a no-change control. These capture permanent changes such as reduced actuator effectiveness or damage, and recurring changes such as adding and later removing a payload. The gain reduction lowers actuator gain to $g=0.5$, actuator damage sets one actuator's gain to zero, and the recurring condition alternates between the original and reduced-gain dynamics every 20k frames after the first change, so each regime lasts 20k frames.

\begin{table*}[t]
\centering
\caption{Walker return after the shift and age--staleness AUC (five seeds, mean $\pm$ s.e.).\\Recency helps after permanent changes but hurts under recurring dynamics and with no change.}
\label{tab:core}
\begin{tabular}{@{}lccccc@{}}
\toprule
Change & $\corr$ & Passive & Recency & Graded & WMAR \\
\midrule
Gain reduction & 1.00 & $177\,\pm\,21$ & $323\,\pm\,30$ & $163\,\pm\,13$ & $387\,\pm\,27$ \\
Actuator damage & 1.00 & $232\,\pm\,40$ & $404\,\pm\,34$ & $246\,\pm\,48$ & $418\,\pm\,28$ \\
Recurring & 0.54 & $481\,\pm\,17$ & $290\,\pm\,19$ & $464\,\pm\,11$ & $484\,\pm\,24$ \\
No change & -- & $936\,\pm\,3$ & $735\,\pm\,27$ & $936\,\pm\,4$ & $816\,\pm\,22$ \\
\bottomrule
\end{tabular}
\end{table*}

\begin{figure*}[t]
\centering
\includegraphics[width=\textwidth]{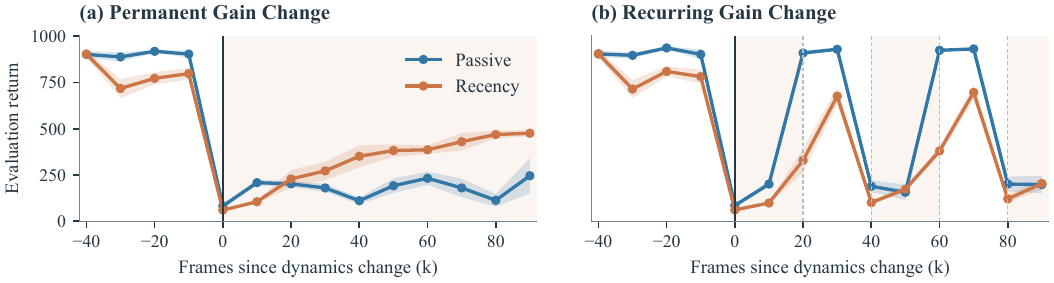}
\caption{Evaluation return before and after the dynamics change for (a) permanent and (b) recurring gain reduction, averaged over five seeds with s.e. bands. The recency window is applied from the branch point, 40k frames before the change. The vertical line marks the initial change, and the shaded intervals mark the periods in which the reduced gain is active. After a permanent change, Recency reaches a higher return than Passive, while under recurrence Passive recovers more strongly whenever the original dynamics return.}
\label{fig:curves}
\end{figure*}

We use separate sweeps to test how recurrence and change magnitude affect whether forgetting older replay helps or hurts performance. The recurrence sweep varies how often the original dynamics return at a fixed gain reduction. The magnitude sweep varies $g$ under a permanent change to identify when Recency begins to outperform Passive. It uses $g\in\{1.0,0.9,0.7,0.5,0.3\}$ and an additional run at $g=0.6$ near the observed sign change.

All experiments use five paired seeds. To control for differences in training history before the shift, all replay strategies within a seed branch from the same checkpoint saved before the shift~\cite{henderson2018matters}. The window size, capacity, and $g=0.6$ controls use a separate set of training runs initialized from independently trained pre-shift checkpoints. We report means and standard errors for all results.

\subsection{Controls, Transfer, and External Perturbations}
The main recency window contains 10k transitions. We repeat the recurrence experiment with a 20k window to test the dependence on window size in \eqref{eq:risk}. We also compare Recency with a 10k reservoir buffer, which keeps a uniform sample from the full replay history rather than only the newest transitions~\cite{vitter1985reservoir,isele2018selective}. Recency and reservoir therefore have the same capacity but different age distributions, allowing us to test whether any benefit comes from keeping recent data specifically rather than simply using a smaller buffer. We also compare against graded replay, which downweights transitions according to the current model's prediction error, to test whether prediction error alone is enough to identify stale data. We include a WMAR-style diversity replay baseline~\cite{yang2024wmar} to test whether preserving a broader range of past experience avoids the costs of strict recency. Finally, we use a reset-once control that clears replay only at the initial change to test whether one-time forgetting is enough, or whether older data needs to be continually discarded.

To test transfer across morphology, we repeat the actuator damage and recurring conditions on Cheetah. We then repeat the same comparison with TD-MPC2 to test a different model-based RL algorithm.

To test whether the replay trends extend beyond our controlled gain changes, we also evaluate two perturbations from the Real-World RL benchmark~\cite{dulacarnold2020rwrl}. Walker joint damping follows a cyclic schedule. Contact friction decreases from $0.7$ toward $0.01$. We do not assign age--staleness AUC values to them because we have not derived validated staleness labels from their perturbation histories.

\textit{Estimator evaluation:} To test whether interaction data can predict when older replay should be discarded without access to ground-truth staleness labels, we evaluate the actuator-response estimator from Section~\ref{sec:estimator} on saved trajectories. For the permanent gain change, we measure detection delay. We then evaluate the selector on 24 saved replay histories, one per training run, spanning three recurrence periods and permanent changes at gain factors $0.5$ and $0.6$. We compare estimated staleness with the ground-truth label at each endpoint and check whether the selector chooses the strategy that performed better. The runs with a 20k recurrence period use the same 314-episode cutoff as the return sweep over recurrence periods in Section VI-B.

\begin{figure*}[t]
\centering
\includegraphics[width=\textwidth]{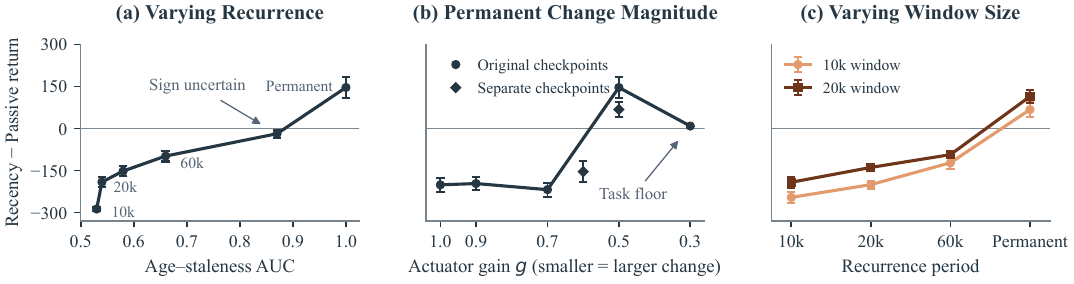}
\caption{Recency--Passive return difference under (a) varying recurrence, (b) varying permanent gain reduction, and (c) varying window size, with positive values favoring Recency. Points show paired means over five seeds with s.e. bars. Conditions shown with diamond-shaped markers were rerun from a separate set of independently trained pre-shift checkpoints.}
\label{fig:twofactor}
\end{figure*}

\textit{Implementation details:} We use an open-source PyTorch implementation of DreamerV3~\cite{nm512dreamerv3torch} with its default proprioceptive DMC configuration. Training uses batches of 16 sequences of length 64, a train ratio of 512, and a model learning rate of $10^{-4}$. Replay capacity is one million transitions. We evaluate for ten episodes every 10k frames, with evaluation trajectories kept separate from training replay, and average return over the 90k frames following the change.

\section{Results}
\label{sec:results}
\subsection{Permanent vs.\ Recurring Dynamics}
Recency helps under permanent changes and hurts under recurrence (Table~\ref{tab:core}, Fig.~\ref{fig:rollouts}). After the permanent gain reduction, Recency improves return over Passive replay by $146\pm38$. Permanent actuator damage gives the same sign, with Recency improving return over Passive by $172\pm42$. In both cases, the measured age--staleness AUC is $1.00$, meaning every stale transition is older than every fresh transition in the buffer.

The direction reverses under recurrence. At a time-averaged AUC of $0.54$, Recency's return is $191\pm17$ below Passive. With no dynamics change, it is $201\pm25$ below Passive. Fig.~\ref{fig:curves} shows how the two replay strategies diverge over training, with Recency already below Passive before the change once its replay window is applied. Graded replay stays close to Passive under recurrence and avoids the large loss from the fixed recency window. WMAR obtains the highest mean return of the four strategies in the Walker gain reduction condition (Table I).

\subsection{Recurrence, Magnitude, and Window Size}
\label{sec:sweep}
We first fix the gain reduction and vary the recurrence period. As the dynamics return more frequently, the time-averaged age--staleness AUC falls from $1.00$ to $0.53$, while the Recency advantage moves from $+146$ to $-287$ (Fig.~\ref{fig:twofactor}a). At the sampled AUC of $0.87$, the Recency advantage is $-19\pm14$, making the sign uncertain. Changing the recurrence period affects both age--staleness AUC and how long the agent trains under each regime. The resulting return differences therefore reflect both factors rather than age--staleness AUC alone.

\begin{figure*}[t]
\centering
\includegraphics[width=\textwidth]{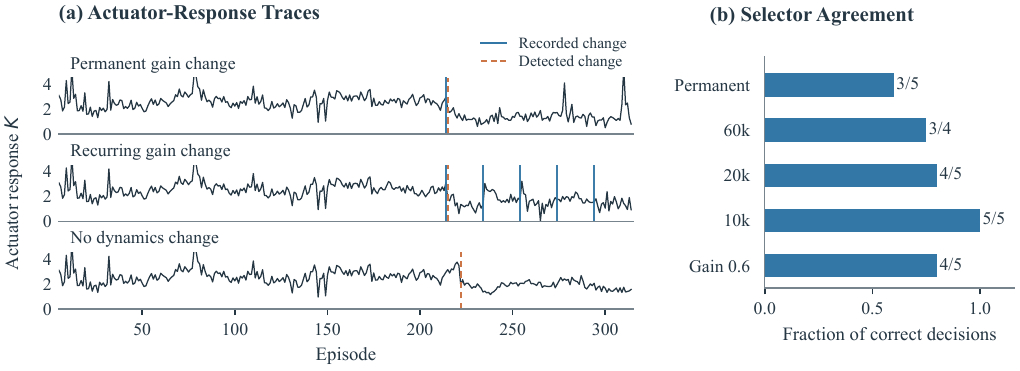}
\caption{Retrospective replay decisions. (a) Actuator-response traces with recorded and detected dynamics changes for permanent, recurring, and unchanged dynamics. The first five random-action episodes are omitted from the plot but included in detection. (b) Agreement between the selector and the better-performing strategy across 24 saved replay histories. Fractions at the end of each bar give the number of correct decisions.}
\label{fig:estimator-validation}
\end{figure*}

The magnitude sweep matches the prediction of \eqref{eq:mstar}: small permanent changes do not justify discarding older data, while larger changes can. The Recency advantage is negative at $g=1.0$, $0.9$, and $0.7$ and becomes positive at $g=0.5$ (Fig.~\ref{fig:twofactor}b). At $g=0.3$, all methods are near the task floor, so the comparison is less informative. The runs from separately trained checkpoints narrow the transition between harmful and useful forgetting to between $g=0.6$ and $g=0.5$. At $g=0.6$, the Recency advantage is $\FreshMild$, while at $g=0.5$ it is $\FreshGain$. The estimator recovers this magnitude from actuator response alone, with mean estimates of $0.43$ at $g=0.5$ and $0.35$ at $g=0.6$, so the quantity that separates these two cases is available to the selector's magnitude gate.

Under recurrence, increasing the replay window reduces Recency's loss relative to Passive at every sampled period (Fig.~\ref{fig:twofactor}c), consistent with \eqref{eq:risk}. A larger window keeps more data from earlier regimes, which becomes useful when those dynamics return. For the replay decision, this means that when previously seen dynamics are likely to recur, retaining older data or widening the window is the better choice, because the transitions being discarded will be needed again.

\begin{table}[t]
\centering
\caption{Walker replay controls, separately trained checkpoints\\(five seeds, mean $\pm$ s.e.).}
\label{tab:capacity}
\renewcommand{\arraystretch}{1.1}
\begin{tabular*}{\columnwidth}{@{\extracolsep{\fill}}lcccc@{}}
\toprule
Change & Passive & Recency & Reservoir & Reset-once \\
\midrule
Gain reduction & $220\,\pm\,21$ & $288\,\pm\,22$ & $128\,\pm\,14$ & $425\,\pm\,66$ \\
Recurring & $487\,\pm\,19$ & $287\,\pm\,12$ & $249\,\pm\,14$ & $312\,\pm\,38$ \\
\bottomrule
\end{tabular*}
\vspace{1.2em}
\caption{Replay comparisons in additional settings\\(five seeds, mean $\pm$ s.e.).}
\label{tab:generality}
\renewcommand{\arraystretch}{1.1}
\begin{tabular*}{\columnwidth}{@{\extracolsep{\fill}}llcc@{}}
\toprule
Setting & Change & Recency $-$ Passive & WMAR $-$ Passive \\
\midrule
Cheetah & Damage & $+184\,\pm\,35$ & $+111\,\pm\,39$ \\
Cheetah & Recurring & $-126\,\pm\,12$ & $-42\,\pm\,22$ \\
TD-MPC2 & Damage & $+8\,\pm\,15$ & -- \\
TD-MPC2 & Recurring & $-123\,\pm\,13$ & -- \\
RWRL & Recurring & $-222\,\pm\,32$ & -- \\
RWRL & Friction & $+94\,\pm\,58$ & -- \\
\bottomrule
\end{tabular*}
\end{table}

\subsection{Replay Controls and Transfer}
Table~\ref{tab:capacity} tests whether Recency helps because it retains recent data or simply because it trains on fewer transitions. Recency and the reservoir buffer both retain 10k transitions, but the reservoir keeps a uniform sample of the full replay history rather than only the newest data. The 10k reservoir performs worse than Passive replay under both permanent and recurring dynamics, while Recency improves return after the permanent gain change. This shows that reducing replay capacity alone does not explain Recency's benefit. Which transitions are retained also matters. Reset-once outperforms Recency after the permanent shift but remains $\ResetRecurringGap$ below Passive under recurrence, and all five of its paired differences there are negative. The recurrence loss persists even without a continually applied window, so discarding data from dynamics that later return is itself costly.

Cheetah reproduces both Walker results (Table~\ref{tab:generality}). On Cheetah, Recency improves return over Passive by $184\pm35$ after actuator damage and is $126\pm12$ below Passive under recurrence. WMAR preserves a broader mix of past experience instead of selecting primarily by age. On Cheetah, it improves less than Recency after damage but avoids part of Recency's loss under recurrence. This supports the main result that aggressive age-based forgetting helps most after permanent shifts, while retaining a broader history is more useful when earlier dynamics can return.

TD-MPC2 shows the same cost under recurrence, with Recency $123\pm13$ below Passive, but a weaker benefit after actuator damage, where the Recency advantage is $+8\pm15$ with only three of five seeds positive. The benefit of forgetting after a permanent shift appears to depend more on the learning algorithm than the cost of discarding data under recurrence. One reason may be that TD-MPC2's Passive return after damage stays near 800, against 232 for DreamerV3 in Table~\ref{tab:core}, leaving little room for Recency to improve on the full history.

On the Real-World RL benchmark, under cyclic damping, Recency is $222\pm32$ below Passive, with all five paired differences negative, extending our recurrence result to an external benchmark perturbation. Under drifting friction, the Recency advantage is $+94\pm58$, but its paired bootstrap interval includes zero, so we cannot conclude that it helps. Across these settings, the cost of forgetting under recurrence is the more consistent effect, and its benefit under changes that do not recur depends on the condition.

\subsection{Retrospective Replay Decisions}
The per-joint actuator-response detector finds the permanent gain change within three episodes in all five seeds. Each detection occurs one episode after the recorded change. The selector chooses Recency in three of the five runs with a permanent change at $g=0.5$ and, across all 24 histories, chooses the replay strategy that actually performed better in 19 of them (Fig.~\ref{fig:estimator-validation}b).

Estimated AUC does not recover every recurring endpoint accurately. Some correct decisions to retain older data instead come from the estimated magnitude falling below the magnitude threshold. The permanent change is detected reliably, while later recurring switches are not always recovered, and the actuator-response estimate can drift over time even when the underlying dynamics do not change.

\section{Discussion}
\subsection{Implications for Replay Design}
A fixed window of recent data can hurt performance even without a dynamics change, because it discards useful older experience and reduces the available training data. Recency's return is $201\pm25$ below Passive with no dynamics change and $191\pm17$ below under recurrence. Large permanent changes can make discarding older data worthwhile, while recurrence makes those transitions useful again. Replay choice therefore depends on both change magnitude and how well transition age separates stale from current data. A larger window reduces Recency's loss under recurrence, while a reservoir with the same capacity, which keeps a uniform sample of the replay history, does not reproduce Recency's gain after a permanent change.

\subsection{From Regime Detection to Replay Selection}
Detecting a change does not determine what to do with older replay data. The actuator-response detector identifies a shift in the interaction statistics. The replay decision also depends on the estimated change magnitude and on how well age separates stale from current data. The results for the permanent change in Fig.~\ref{fig:estimator-validation} make this distinction clear. The gain change is detected in every seed one episode after it occurs, yet the selector chooses the better-performing strategy in only three of the five runs.

In both runs with a wrong decision, the change is still detected at the same delay as in the successful seeds. In one seed, the estimated AUC is 0.72 and the estimated magnitude is 0.36, so neither gate is passed. In the other, the magnitude estimate passes at 0.43, but the estimated AUC falls to 0.47. In these cases, detecting the change is easier than deciding whether older replay should be discarded.

\textit{Limitations:} Age--staleness AUC summarizes how well age ranks stale and fresh transitions, but it does not specify how much stale data falls inside a particular replay window. In the controlled experiments, staleness is defined from known dynamics regimes and reconstructed episode timelines. We do not report AUC for the Real-World RL perturbations because validated staleness labels have not been derived from those perturbation histories.

The actuator-response analysis is retrospective, and its thresholds and the histories it is evaluated on both come from Walker. The actuator-response estimate can also vary with policy behavior and state visitation. An online version would need to make each decision from past observations only, for example with online change-point detection~\cite{adams2007bocpd}, and be evaluated on previously unseen conditions.

\section{Conclusion}
We show that the value of older replay data depends on how the dynamics evolve over time. Restricting replay to recent data improves return after large permanent changes, while older data becomes useful again when the dynamics recur. Recency's loss under recurrence also appears on Cheetah, with TD-MPC2, and under an external benchmark perturbation. Our retrospective actuator-response estimator suggests that these replay decisions can be made from interaction data without access to simulator dynamics parameters. An online version could use the same signals to adjust replay retention as a robot's dynamics change over time.

\section*{Acknowledgments}
We thank Rohan Bandaru, Brian Wei, Sergio Orozco, David Paulius, and members of the Brown Intelligent Robot Lab for helpful discussions and feedback.

\bibliographystyle{IEEEtran}
\bibliography{references}
\end{document}